# Handwriting Trajectory Recovery using End-to-End Deep Encoder-Decoder Network

[a]Ayan Kumar Bhunia, [a]Abir Bhowmick, [b]Ankan Kumar Bhunia, [a]Aishik Konwer, [c]Prithaj Banerjee, [d]Partha Pratim Roy[*], [e]Umapada Pal

[a]Department of ECE, Institute of Engineering & Management, Kolkata, India
[b]Department of EE, Jadavpur University, Kolkata, India
[c]Department of CSE, Institute of Engineering & Management, Kolkata
[d]Departmentof CSE, Indian Institute of Technology Roorkee, India
[e]CVPR Unit, Indian Statistical Institute, Kolkata, India
*email: 2partharoy@gmail.com

*Abstract*—In this paper, we introduce a novel technique to recover the pen trajectory of offline characters which is a crucial step for handwritten character recognition. Generally, online acquisition approach has more advantage than its offline counterpart as the online technique keeps track of the pen movement. Hence, pen tip trajectory retrieval from offline text can bridge the gap between online and offline methods. Our proposed framework employs sequence to sequence model which consists of an encoder-decoder LSTM module. The proposed encoder module consists of Convolutional LSTM network, which takes an offline character image as the input and encodes the feature sequence to a hidden representation. The output of the encoder is fed to a decoder LSTM and we get the successive coordinate points from every time step of the decoder LSTM. Although the sequence to sequence model is a popular paradigm in various computer vision and language translation tasks, the main contribution of our work lies in designing an end-to-end network for a decade old popular problem in document image analysis community. Tamil, Telugu and Devanagari characters of LIPI Toolkit dataset are used for our experiments. Our proposed method has achieved superior performance compared to the other conventional approaches.

## I. Introduction

Automatic handwriting analysis and recognition has been an extensive field of research interest by the scientists since the last two decades [26]. Based on acquisition technique, methods for handwritten character recognition can be classified into two categories: Offline acquisition and Online acquisition. Online method requires special input devices such as electrostatic or electromagnetic tablets and stylus. On the other hand, in case of offline handwritten character recognition, a character written on a paper is captured by a scanner or a camera at a high resolution. Then it is stored as a digital image.

Compared to online method, offline one is much more difficult because of the limited availability of information. A static two-dimensional (2D) image captured by scanning handwritten characters on a paper is the only available information here. Dynamic information captured by tablet includes velocity, pressure, inclination etc. But an image scanner or video camera is not capable of extracting this kind of information. That is why the offline method has to undergo some complicated image processing methods to analyze handwritten character segments for feature extraction. In the online method, the coordinate information of the pen-tip trajectory is available. This is stored as a one dimensional (1D) vector. On the contrary, the temporal information is unavailable in case of offline method. Therefore, if we are able to recover the stroke trajectory from the static 2D image, the offline recognition problem can be viewed as the online one. Thus, extraction of the dynamic information is a connection between the offline method and its online counterpart. Examples of offline and online images are shown in Fig. 1.

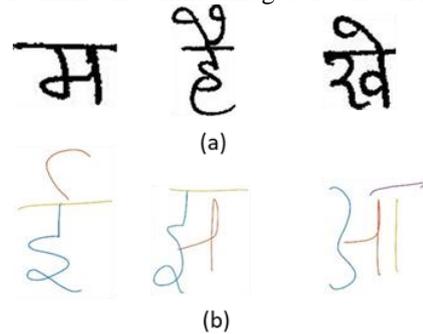

Fig.1. Example showing (a) Offline and (b) Online Devanagari characters. Different colors in (b) indicate different stroke in online example.

In this paper, we propose a novel technique that can predict the probable trajectory of an offline character level image. A deep convolutional neural network is used to extract a sequence of features from the handwritten image. The objective of our model is to convert this sequence to another sequence that gives the 2D coordinate points of the pen trajectory. However, simple recurrent neural network cannot solve this problem as the length of the input and output sequence may not match. An encoder-decoder LSTM network is able to solve this problem by introducing two networks; an encoder network that encodes the feature sequence to a hidden representation and a decoder network that takes the representation and sequentially predicts the data points. The goal of our proposed architecture is to retrieve the stroke trajectory or drawing order of an offline character which can be seen as a conversion of two-dimensional image matrix as a one-dimensional sequence which consists of successive coordinate positions of the pen-tip. In our experiment, we have considered character level data for stroke recovery. In future, we plan to extend this architecture for word level data as well as general trajectory recovery from any hand drawn sketch images.

The contributions of our work are as follows: Firstly, we develop an end-to-end deep architecture for handwriting stroke recovery task, which can be trained in an end-to-end manner. Over the past two decades, this stroke recovery problem has been addressed using some local handcrafted features and heuristic assumptions. Our work is the first to introduce deep learning in this decade old problem. Secondly, we have done extensive experiments on Tamil, Telugu and Devanagari characters of LIPI Toolkit dataset to show the performance of our proposed architecture. In Section II, we mention some earlier approaches for stroke recovery. We describe our proposed architecture and implementation details in Section III and IV respectively. In Section V, we detail the experimental results. Section VI deals with the conclusion and future works.

## II  Related Work

Researches on pen trajectory retrieval technique have been carried out with the help of two major processes: analysis of local features and global features. Local feature analysis [5] includes the information which can be extracted from a local region of the static image like line segment, junction point, ambiguous zone, endpoints etc. All these features help in analyzing the global characteristics of the image. In global feature analysis phase [18], the overall image is examined using the data obtained from the local phase.

Lee and Pan [1, 2] traced offline signature using hierarchical decision making for stroke identification and ordering based on a set of heuristic rules. Critical point normalization and segmentation were used to make the algorithm independent of rotation, translation and scaling. This method transforms a two-dimensional spatial pattern into a one-dimensional temporal pattern. Govindaraju and Krishnamurthy [4] used holistic recognition of offline cursive characters using the temporal information extracted from the online one. Doermann and Rosenfeld [5, 7] proposed a taxonomy of local, global and regional clues to recover temporal information of offline handwriting. Precise position and direction of stylus or pen-tip as a function of time, pen-up, pen-down points and a sequence of intermediate points, pressure, velocity, acceleration - this kind of temporal information is available here. Boccignone et al. [8] implemented a good continuity criteria which include direction, length etc. of the character to extract the dynamic information. Abuhaiba et al. [11] proposed three algorithms for offline Arabic handwriting recognition. The first algorithm is used to convert a smoothed image to a polygonal estimation. The second algorithm gives the starting point of the handwritten character and the third algorithm makes use of graph traversal technique in order to extract the temporal information of an Arabic character. Privitera et al. [12] provided a segmentation method to recover the motor-temporal information from a handwritten word. The work in [14] classified 150 different words written by four different writers with the help of the state-of-the-art online method. In [15], Rousseau et al. examined the intrinsic knowledge of handwritten characters to analyze its usefulness in trajectory retrieval. In some experiment on signatures [25], information obtained earlier from the on-line technique has been used to recover off-line drawing order.

Viard-Gaudin et al. [18] analyzed offline cursive handwriting using HMM. In the first approach, a feature vector is extracted scanning the image in left to right direction. In the second approach, a set of paths or edges is generated using a graph model that contains the probable temporal information. Lallican and Viard-Gaudin [16] proposed a method where they used grey level information of the image and a Kalman filtering approach that can predict the probable stroke trajectory of the offline character. Qiao et al. [24] employed a matching cost function to recover drawing order of a multistroke character with the help of best-matched paths from a set of strokes. It can effectively minimize the searching space and can remove the dependency on the start and the end vertices. A research on Chinese handwritten character recognition has been carried out by Liu et al. [17]. Kato and Yasuhara [22, 23] recovered both the single stroke and multi stroke character drawing order with the help of graph theory. Nel and du Preez [20, 21] employed second order HMM to estimate the stroke trajectory. The main challenging parts of trajectory retrieval are to find out the correct starting point and extract the correct incoming and outgoing paths from junction points as temporal information is unavailable in case of the offline images. Our proposed method handles these issues effectively.

## III  Proposed Framework

In this section, we detail our proposed offline to online data translation process. The model can be trained end-to-end to recover the online information from offline handwritten text images. Our network is inspired from the recent sequence to sequence model [6, 10] based on encoder-decoder architecture. The model sequentially predicts the data coordinate points of the pen trajectory from the offline character level images. It involves mainly two steps; first we extract a sequence feature vector from the offline images using convolutional neural network. Next, an encoder-decoder LSTM network outputs the required coordinate points. Extraction of feature sequence vector is a major step for our architecture. Widely used popular hand crafted feature extraction processes like HOG, SIFT cannot acquire a problem

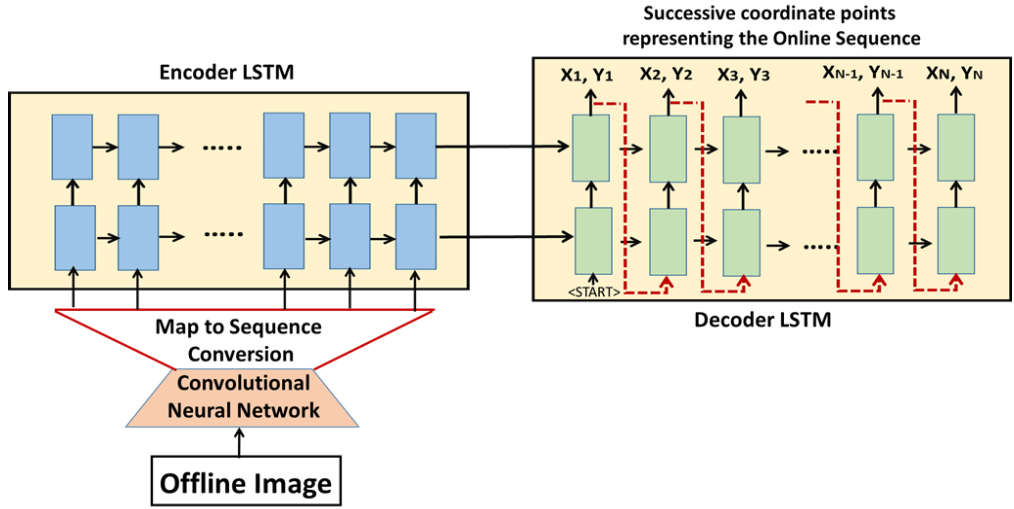

Fig.2. Diagram showing proposed encoder-decoder architecture for handwriting stroke recovery problem. For sake of simplicity, we have shown unidirectional LSTM in this diagram

specific feature representation. Due to the recent success of convolutional neural network (CNN) to capture robust feature representation, we employ a deep CNN network to learn a feature sequence directly from the images. Handwritten character images of size 64x64 are the input of the CNN network. The network is composed of the convolutions and the max pooling operations removing the fully connected layers. Six convolutional layers with kernel size 3x3 are stacked to learn a deep feature representation. We adopt 1x2 pooling method [13] to yield longer sequence vector. As we are trying to predict the set of many trajectory points from our model, it is helpful to have a longer sequence vector to enhance the performance of the LSTM encoder-decoder network. Each feature vector in the output sequence of convolutional network is considered as an image descriptor of a particular rectangular portion of the image. However, dealing with the sequence and converting it to another sequence is a challenging task. A recurrent neural network is useful to convert a sequence vector to another sequence vector of same length. The RNN unit iteratively updates the hidden state value $h_t$ by a non-linear function $h_t = g(x_t, h_{t-1})$. Given a sequence of inputs $x_1, x_2, x_3, ..., x_n$, a basic RNN network returns an output sequence $y_1, y_2, y_3, ..., y_n$ using the following equations:

$$h_t = sigmoid(U.x_t + V.h_{t-1}) \quad \ldots \ldots (1)$$

$$y_t = W.h_t \quad \ldots \ldots (2)$$

where, $U, V$ and $W$ are the learning parameters. But the basic RNN fails to output a sequence of different length. It means that we cannot apply the basic RNN network to predict the coordinate points from the extracted sequence vector in our work. The Sequence to sequence learning model [6, 10] overcomes this problem by using recurrent encoder-decoder architecture. The basic idea of this model is to input the extracted sequence vector to a network called Encoder that encodes it to a high dimensional feature representation.

Another network called Decoder is used to extract the output sequence which are the data coordinate points in our case. The encoder and decoder networks are recurrent neural network (RNN) where the second network is conditioned on the input sequence. As the basic RNN unit suffers from the vanishing gradient problem, LSTM network is used which is specially designed to learn longer dependencies. The internal gate structure of the LSTM network allows it to learn long temporal dependencies. Although, the LSTM network eliminates the vanishing gradient problem, the basic unidirectional LSTM may not work well as it captures sequential information from left to right direction only. Thus bidirectional LSTM network [9] is adopted to learn dependencies in both directions.

Given an input sequence $X = x_1, x_2, x_3, ..., x_n$ and a target sequence $Y = y_1, y_2, y_3, ..., y_{n'}$ the seq2seq model learns to predict the conditional probability $p(y_1, y_2, y_3, ..., y_{n'}|x_1, x_2, x_3, ..., x_n)$. The Encoder network first computes the encoded feature representation $v$ of fixed dimension from the extracted sequence X. We get the encoded output from the last hidden state of the LSTM unit. Then, the encoded output $v$ is processed in the Decoder network to output the final sequence of coordinate points. The hidden state of the Decoder network is initialized by the encoder output $v$. The output of the previous time-step is the input at the next time-step of the decoder. The final conditional probability distribution is obtained:

$$p(y_1, y_2, ..., y_{n'}|x_1, x_2, ..., x_n; \theta)$$
$$= \prod_{t=1}^{n'} p(y_t|v, y_1, y_2, ..., y_{t-1}; \theta) \quad \ldots \ldots (3)$$

We can obtain each $p(y_t|v, y_1, y_2, ..., y_{t-1})$ from the decoder network at a particular time step $t$ by applying a fully connected layer over the decoder representation $u_t$.

$$p(y_t|v, y_1, y_2, ..., y_{t-1}; \theta) = W_o.u_t + b_o \quad \ldots \ldots (4)$$

where, the dimension of $W_o$ and $b_o$ is such that we get 2D data points from each time-step of the decoder network. The model continuously outputs the predicted coordinate data points. We use L1 distance loss function as objective function for our model. We obtain the loss by computing the L1 distance for every predicted data point $\vec{z_t}$ with their corresponding ground truth points $\vec{Z_t}$.

$$L = \frac{1}{n'} \sum_{t=1}^{n'} \left\| \vec{z_t} - \vec{Z_t} \right\|_1 \qquad \ldots\ldots (5)$$

The overall architecture is shown in Fig.2.

## IV. IMPLEMENTATION DETAILS

The feature sequence generator is a CNN network of six convolutional layers. It takes input images of size 64x64. We use the kernel size of 3x3 for all the layers. We adopt 1x2 pooling method as described in [13] instead of a traditional pooling technique. Rectified linear unit (RELU) is used after each convolutional layer. Batch normalization is applied after $3^{rd}$ and $4^{th}$ convolutional layers to speed up the training process. Also, L2 regularization is applied to the weights of the networks. The encoder and decoder network of the seq2seq model is parameterized by a bidirectional LSTM network. Two bidirectional LSTM cells of 512 hidden units are stacked to have higher abstraction ability. Experiments are conducted on a server with Nvidia Titan X GPU with 12 GB of memory. We implemented the model using Tensorflow. The model is optimized by Adam optimizer with learning rate 0.001. We train the model for 200 epochs with batch size 32. It takes around 5-6 hours for training in the Nvidia Titan X GPU system.

Since we are using simple L1 distance loss to train our network, the decoder outputs continuous values for coordinate points which may deviate from the actual skeleton of the offline images. We translate these predicted coordinate points to the nearest point on the skeleton of the offline image as a post processing step.

## V. EXPERIMENT & RESULT ANALYSIS

**Datasets**: Deep learning based framework requires a large amount of training data for better generalization. To the best of our knowledge, there exists no such dataset with large amount of data containing trajectory information of offline images. However, a point is to be noted that pre-recorded offline images do not contain any time distributed pen's trajectory information, unless the online coordinate points are recorded sequentially at the time of writing. Hence, we use a reverse approach to generate our dataset. Real online data contains trajectory information in terms of successive coordinate points. We convert the online data to its offline equivalent by redrawing the image using the online coordinate points. For these converted offline images, we have the online coordinate points representing online trajectory movement of the pen. We use this data to train our encoder-decoder based sequence to sequence model for our experiment. We perform morphological thickening operation on the converted offline images to make it similar to original offline images. Few examples of converted offline images are shown in Fig.- 3. From the sequence of online coordinate points of each data, 50 points are sampled uniformly over the complete trajectory to make it uniform. Following this, we normalize these coordinate values within 64x64 range in accordance with the offline character images which are also resized to a size of 64x64 unit.

We have used LIPI Toolkit dataset[1] in our experiments, which contains character level online data from three Indic scripts, namely Tamil, Telugu and Devanagari. Complex shape, multiple junction points and irregular starting points while tracing from one character to another are the common characteristics of Tamil, Telugu and Devanagari scripts. Because of the above variations, we choose these 3 Indic scripts for experimentation. Table I shows the number of sample considered for training, testing and validation for three scripts respectively.

**Table I: Dataset details**

| Scrips | Training | Testing | Validation |
|---|---|---|---|
| Tamil | 20,789 | 8,789 | 4,897 |
| Telugu | 21,485 | 8,687 | 4,856 |
| Devanagari | 20,987 | 8,239 | 4,789 |

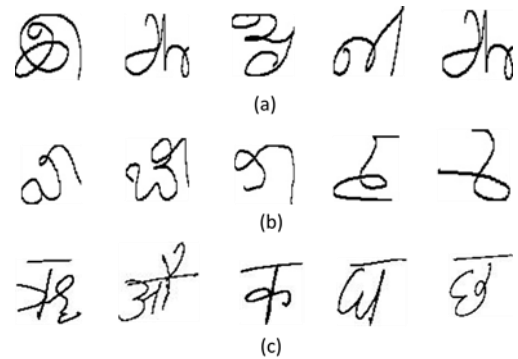

Fig 3: Example showing (a) Tamil (b) Telugu and (c) Devanagari offline character images (converted from real online data). These offline images are fed into our network to predict the sequence of coordinate points representing the pen's trajectory.

**Evaluation Metrics:** We measured our accuracy based on these evaluation metrics-

1) Starting Point (SP) accuracy: If the network can predict the correct starting point of the offline image, we count it as a positive result, otherwise negative. For every offline image, there is only one starting point, hence we calculate the starting point detection accuracy as follows:

$Starting\ Point\ Accuracy = \frac{Number\ of\ images\ with\ correct\ SP\ detection}{Total\ number\ of\ test\ images}$ …. (6)

2) Junction Points (JP) accuracy: While traversing a junction point, if it can correctly determine the incoming and outgoing path of the junction point, we consider it as positive junction point traversal. One offline character may have more than one

---
[1] lipitk.sourceforge.net/hpl-datasets.htm

junction points. Hence, we define junction point accuracy as follows:

$$Junction\ Point\ Accuracy = \frac{Number\ of\ JP\ correctly\ traversed}{Total\ number\ Junction\ points\ in\ test\ data} \quad ....(7)$$

3) Complete Trajectory (CT) retrieval accuracy: One offline character image may have multiple junction points. It may happen that it traverses one junction point perfectly but fails for others. Hence, we evaluate this metric as a positive result when complete trajectory of an offline character image is perfectly retrieved along with the correct starting point.

**Results:** We have evaluated the performance of handwriting trajectory recovery problem using three different evaluation metrics. For each script, we have trained individual model and the stroke recovery results are reported in Table II. We have compared the accuracy of our proposed method with one traditional framework [22] for the same dataset used in our work. Kato et al. [22] employed a graph traversal based basic trace algorithm which can trace a single stroke handwritten character. If a vertex and an edge connected to it are given, it traces the graph until it encounters an odd degree vertex. Despite having a number of research papers in this field, we have chosen this paper based on the popularity (number of citations) and it has also been done on character level. Here, we have used simple L1 distance loss for training and thus the decoder predicts continuous output for coordinate points. Hence, it is expected to look into the predicted coordinate values in more details. In Fig. 4, we have shown the distribution of X and Y coordinate values on the validation dataset with respect to the number of steps of iteration. From this figure, it can be inferred that the coordinate values do not diverge from the 64x64 range, as we normalized the coordinate values and size of the character images to 64x64. A few qualitative results of methods are shown in Fig 5. A few erroneous stroke recovery results are also shown in Fig. 6.

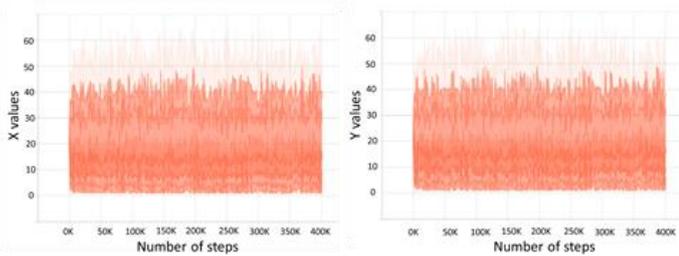

**Fig 4: Distribution of predicted X and Y coordinate values at the output of decoder LSTM. Highly dense colored region signifies that most of the predicted values lie within that range.**

## VI. CONCLUSION

In this work, we have introduced a complete deep learning based solution for handwriting trajectory recovery which has been studied extensively over last two decades in Document Image Analysis community. Our framework is based on sequence to sequence translation model using bidirectional LSTM. We have used a convolution neural network in order to extract sequential features from offline character images. However, our proposed architecture is a basic neural translation model [10] which does not have any attention mechanism [3]. In our future works, we will consider this attention mechanism for better long term dependencies and extend our framework for word level offline images. Also, this architecture can be extended for trajectory retrieval from handwritten sketch images. One of the major limitations is that we need to train separate model for individual script to retrieve handwriting trajectory. As a future research direction, the designing of script independent universal handwriting trajectory retrieval can be addressed.

**Table II: Stroke recovery accuracy**

|  | Evaluation Metrics | Tamil | Telugu | Devanagari |
|---|---|---|---|---|
| Kato et al. [22] | Starting Point (SP) Accuracy | 91.69% | 92.16% | 92.87% |
|  | Junction Points (JP) accuracy | 91.02% | 91.31% | 92.31% |
|  | Complete Trajectory (CT) retrieval accuracy | 88.34% | 89.03% | 90.42% |
| Proposed | Starting Point (SP) Accuracy | 98.12% | 98.01% | 98.57% |
|  | Junction Points (JP) accuracy | 97.04% | 97.14% | 97.69% |
|  | Complete Trajectory (CT) retrieval accuracy | 95.54% | 95.81% | 96.35% |

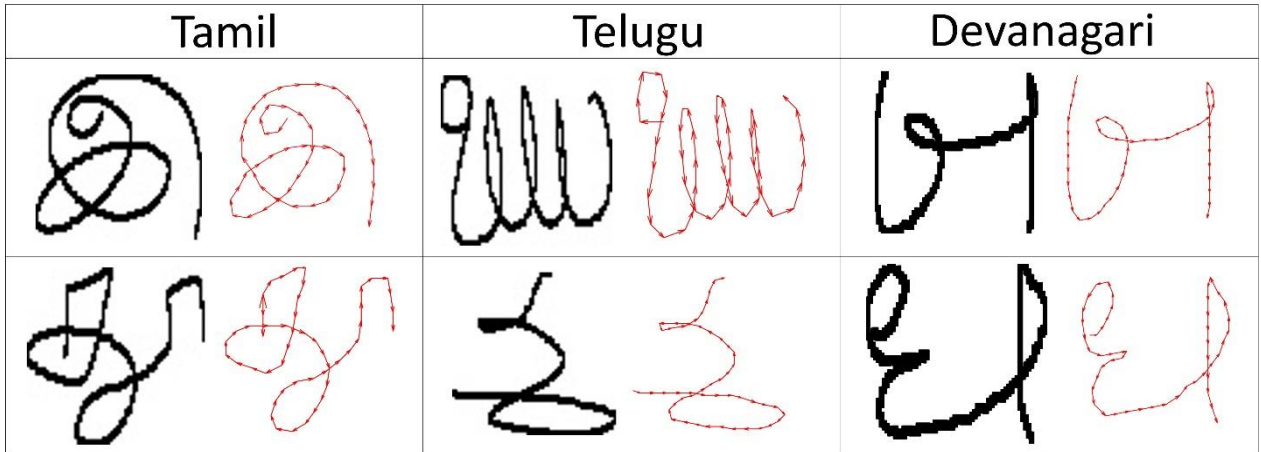

Fig 5: Qualitative results showing stroke order recovery (For directional information, please refer the online version of this document)

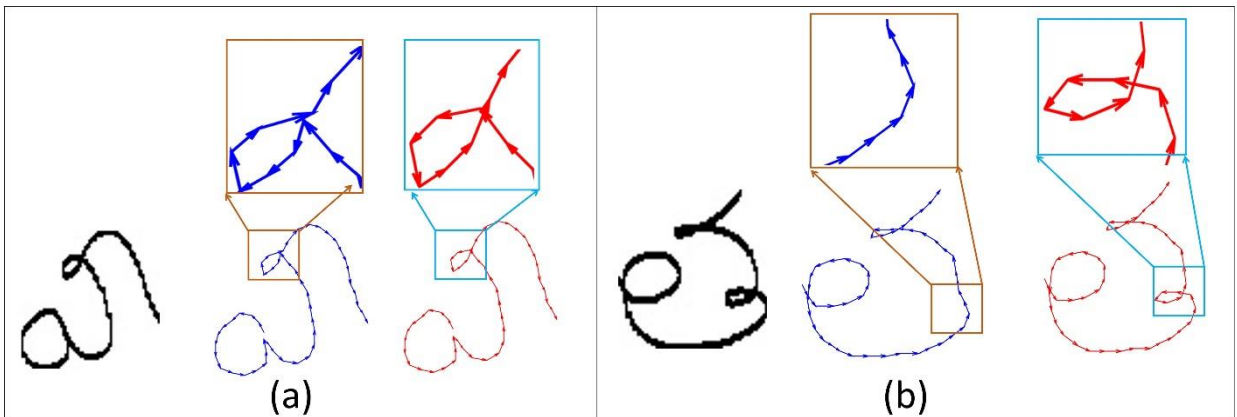

Fig 6: Example showing erroneous trajectory recovery (predicted trajectory in blue and actual ground truth in red). (a) From the junction point, it traverses the loop in wrong direction. (b) It misses one loop during traversal. (For directional information, please refer the online version of this document)